\documentclass[lettersize,journal]{IEEEtran}
\usepackage{amsmath,amsfonts}
\usepackage{algorithmic}
\usepackage{algorithm}
\usepackage{array}
\usepackage[caption=false,font=normalsize,labelfont=sf,textfont=sf]{subfig}
\usepackage{textcomp}
\usepackage{stfloats}
\usepackage{url}
\usepackage{verbatim}
\usepackage{graphicx}
\usepackage{cite}
\usepackage{multirow}
\usepackage{arydshln}
\usepackage{orcidlink}
\begin{document}

\title{Enhancing Tabular Anomaly Detection via Pseudo-Label-Guided Generation}

\author{
Wei Huang\dag~\orcidlink{0000-0003-2262-9339}, Yuxuan Xiong\dag~\orcidlink{0009-0006-5851-3951}, Hezhe Qiao~\orcidlink{0000-0003-3511-0528}, Yu-Ming Shang*~\orcidlink{0000-0003-2903-4223}, Xiangling Fu*~\orcidlink{0000-0002-1492-2829}, Guansong Pang~\orcidlink{0000-0002-9877-2716}~\IEEEmembership{Member,~IEEE}
\thanks{
\dag~These authors contributed equally; *~Corresponding author; This work was supported by the National Natural Science Foundation of China (No.72274022).}
}

\markboth{Journal of \LaTeX\ Class Files,~Vol.~14, No.~8, August~2021}%
{Shell \MakeLowercase{\textit{et al.}}: A Sample Article Using IEEEtran.cls for IEEE Journals}


\maketitle

\begin{abstract}
Identifying anomalous instances in tabular data is essential for improving data reliability and maintaining system stability. 
Due to the scarcity of ground-truth anomaly labels, existing methods mainly rely on unsupervised anomaly detection models, or exploit a small number of labeled anomalies to facilitate detection via sample generation or contrastive learning.
However, unsupervised methods lack sufficient anomaly awareness, while current generation and contrastive approaches tend to compute anomalies globally, overlooking the localized anomaly patterns of tabular features, resulting in suboptimal detection performance.
To address these limitations, we propose PLAG, a pseudo-label-guided anomaly generation method designed to enhance tabular anomaly detection. 
Specifically, by utilizing pseudo-anomalies as guidance signals and decoupling the overall anomaly quantification of a sample into an accumulation of feature-level abnormalities, PLAG not only effectively obviates the need for scarce ground-truth labels but also provides a novel perspective for the model to comprehend localized anomalous signals at a fine-grained level.
Furthermore, a two-stage data selection strategy is proposed, integrating format verification and uncertainty estimation to rigorously filter candidate samples, thereby ensuring the fidelity and diversity of the synthetic anomalies. 
Ultimately, these filtered synthetic anomalies serve as robust discriminative guidance, empowering the model to better separate normal and anomalous instances.
Extensive experiments demonstrate that PLAG achieves state-of-the-art performance against eight representative baselines. 
Moreover, as a flexible framework, it integrates seamlessly with existing unsupervised detectors, consistently boosting F1-scores by 0.08 to 0.21.
\end{abstract}


\section{Introduction}
Tabular data represent a structured data modality organized in a two-dimensional format, where rows denote individual instances and columns correspond to feature attributes. 
As one of the most ubiquitous and representative data types in real-world applications, tabular data serve as the backbone for critical domains such as finance~\cite{al2021financial,park2024enhancing}, healthcare~\cite{fernando2021deep}, and network security~\cite{ahmad2021network}. 
In modern intelligent systems, identifying tabular anomalies is indispensable for enhancing data reliability and safeguarding system stability. 

Given the extreme scarcity of annotated anomalies in real-world scenarios, existing research predominantly focuses on tabular anomaly detection within an unsupervised learning paradigm. 
To circumvent the reliance on anomaly labels, these methods exclusively model the distribution of normal samples during the training phase. 
Subsequently, during inference, test instances that deviate significantly from this learned distribution are identified as anomalies.
For example, OCSVM~\cite{scholkopf1999support} and DeepSVDD~\cite{ruff2018deep} learn a separating hyperplane that places normal samples on one side, and samples that fall on the opposite side during inference are identified as anomalies. 
NeuTraL AD~\cite{qiu2021neural}, ICL~\cite{shenkar2022anomaly} and MCM~\cite{yin2024mcm} design distinct self-supervised tasks to enhance the model's understanding of the feature patterns of normal samples. During testing, samples that yield high self-supervised losses are regarded as anomalies.
While these methods have achieved promising results in anomaly detection, the absence of anomaly guidance often limits their effectiveness in accurately discriminating between normal and anomalous samples~\cite{perini2024unsupervised}.
\begin{figure}
    \centering
    \includegraphics[width=\columnwidth]{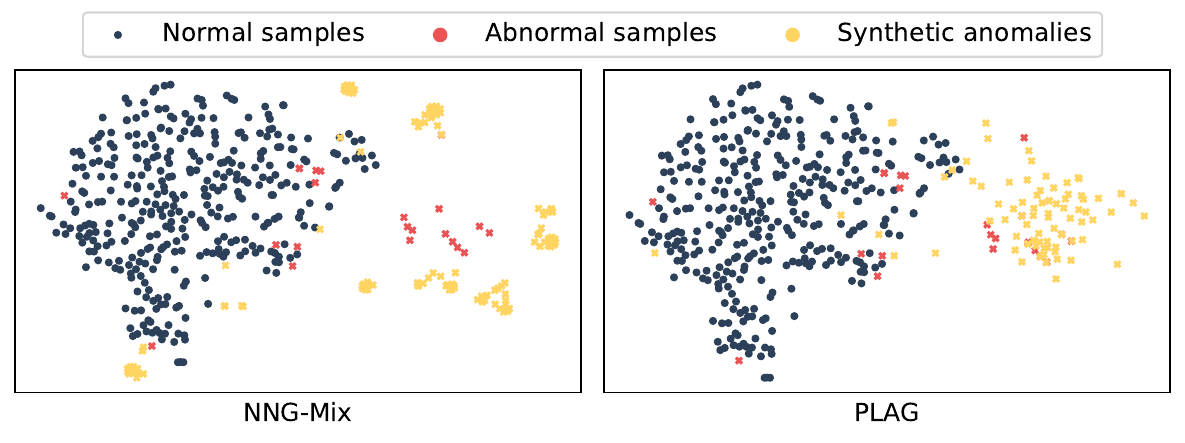}
    \caption{Visualization of Anomaly Generation.
    (a) Distribution of synthetic anomalies generated by replacing the labeled anomalies in NNG-Mix with pseudo anomalies.
    (b) Distribution of synthetic anomalies generated by the proposed method.}
    \label{fig:problem}
    \vspace{-1.5em}
\end{figure}

To alleviate this limitation, some approaches attempt to introduce anomaly prior guidance by synthesizing anomalous samples~\cite{dong2024nng} or generalizing anomaly ranking capabilities from known instances to unlabeled data~\cite{li2023deep}. 
These strategies effectively enhance the model's discriminative ability between normal and anomalous samples, thereby boosting overall detection performance.
Unfortunately, these methods inherently rely on the assistance of labeled anomalies.
Therefore, in the absence of available labels, these approaches are rendered ineffective and fail to deliver performance improvements.
Furthermore, existing methods predominantly measure anomalies directly within a holistic latent embedding space, neglecting the inherent reality that tabular features are highly heterogeneous and implicit~\cite{borisov2022deep,grinsztajn2022tree}. 
Consequently, localized anomalous signals in specific dimensions are frequently submerged by the global representation, making it difficult to provide a fine-grained characterization of anomalous patterns.
As illustrated in Fig.~\ref{fig:problem}~(a), NNG-Mix~\cite{dong2024nng} synthesizes anomalies directly within the global embedding space by relying on nearest neighbor constraints and Gaussian perturbations. Due to the lack of a precise perception of fine-grained features, it tends to yield overly scattered synthetic samples, ultimately failing to faithfully approximate the true anomaly distribution.

To address these issues, we propose \textbf{PLAG}, a \underline{\textbf{P}}seudo-\underline{\textbf{L}}abel-guided \underline{\textbf{A}}nomaly \underline{\textbf{G}}eneration method to enhance tabular anomaly detection.
By leveraging pseudo-labels to direct a large language model (LLM) in capturing the anomalous semantics of tabular data and synthesizing candidate anomalies, the proposed method not only effectively mitigates the real-world scarcity of annotated anomalies, but also provides a novel perspective for the model to comprehend localized anomalous signals at a feature-specific level.
Specifically, to overcome the scarcity of ground-truth anomaly labels, PLAG employs an existing unsupervised anomaly detector to score unlabeled data, assigning pseudo-labels to a small subset of instances exhibiting high anomaly probabilities.
To tackle the issue where global anomaly measurements often obscure localized anomalous signals, PLAG decouples the overall anomaly quantification of a sample into the accumulation of feature-level anomalies. 
By treating each feature dimension as an independent distribution, a sample's comprehensive anomaly score is simplified as the cumulative sum of the rarity of each feature value within its respective distribution. 
This straightforward yet effective mechanism empowers the LLM to precisely capture feature-level anomalous semantics and generate synthetic samples that conform to these underlying patterns.
Furthermore, we propose a two-stage data selection strategy that integrates format verification and uncertainty estimation to filter candidate samples, achieving a balance between the fidelity and diversity of the synthetic anomalies. 
Ultimately, these filtered high-quality synthetic samples, as depicted in Fig.~\ref{fig:problem}~(b), faithfully approximate the true anomaly distribution, thereby providing reliable discriminative guidance for the downstream model.

Extensive experiments on five tabular datasets demonstrate that PLAG outperforms eight various anomaly generation approaches. 
Furthermore, as a flexible framework, PLAG can be seamlessly integrated with existing base models. 
When combined with eight representative unsupervised anomaly detection methods, it yields substantial F1 score improvements ranging from 0.08 to 0.21.
The main contributions of this paper are as follows:
\begin{itemize}
    \item 
    We propose PLAG, a pseudo-label-guided anomaly generation method for tabular data. 
    By introducing pseudo-anomaly signals as prior guidance and decoupling the overall anomaly quantification of a sample into an accumulation of feature-level patterns, PLAG effectively guides a large language model to precisely capture the underlying anomalous semantics and generate high-quality synthetic anomalies accordingly. 
    The proposed method not only breaks through the bottleneck of severe label scarcity in real-world scenarios but also provides a novel generative perspective for resolving localized anomalous signals at a fine-grained level.

    \item 
    We propose a two-stage data selection strategy that integrates format verification and data-driven uncertainty estimation to balance the fidelity and diversity of synthetic samples.
    Specifically, grounded in fuzzy rough set theory, this strategy innovatively quantifies sample uncertainty via concept discernibility rather than traditional geometric proximity. This effectively mitigates the inherent ambiguity of pseudo-labels and the noise introduced during the LLM synthesis process.

    \item Extensive experiments demonstrate that PLAG achieves state-of-the-art anomaly detection performance compared with 8 representative anomaly generation baselines. Furthermore, PLAG serves as a flexible framework for generating anomalies, which can be integrated with various base detectors. When combined with eight existing unsupervised tabular anomaly detection models, PLAG yields consistent performance gains, improving their F1 scores by margins ranging from 0.08 to 0.21.
\end{itemize}

\section{Related Work}
\subsection{Unsupervised Tabular Anomaly Detection}
Since true labels are difficult to obtain in practice, existing tabular anomaly detection methods typically operate under an unsupervised setting. These methods learn the feature distribution of normal samples and classify instances that deviate from the learned distribution as anomalies during inference.
A variety of techniques have been proposed to enhance anomaly detection performance, including density-based analysis~\cite{breunig2000lof,liu2022unsupervised}, data perturbation~\cite{cai2022perturbation,dai2024unsupervised}, distribution modeling~\cite{li2020copod,li2022ecod}, and feature pattern learning~\cite{liu2008isolation,zong2018deep}.
For example, OCSVM~\cite{scholkopf1999support} and DeepSVDD~\cite{ruff2018deep} learn a hyperplane or hypersphere during training that separates all normal samples to one side. During inference, samples falling on the opposite side of the hyperplane or hypersphere are considered anomalies. 
NTL~\cite{qiu2021neural} and ICL~\cite{shenkar2022anomaly} design self-supervised tasks to learn the features of normal samples and identify test samples that perform poorly in these tasks as anomalies. 
MCM~\cite{yin2024mcm} employs a masked autoencoder to reconstruct normal samples with masked features back to their original form, learning the normal pattern and classifying test samples that deviate from this learned pattern as anomalies.
Although these methods have achieved promising results, they rely exclusively on identifying anomalies based on their deviation from a fitted normal distribution. Such a mechanism leads to a vague awareness of actual anomalous patterns, which limits their overall detection effectiveness~\cite{kocak2019safepredict,hullermeier2021aleatoric,perini2024unsupervised}.

\subsection{Tabular Data Generation}
\textbf{Anomaly Generation Methods.}
With respect to tabular data, some studies have introduced various techniques for anomaly sample generation, including template design, feature concatenation, noise addition and distribution learning. These methods primarily construct anomalous samples by modifying the feature representations of known instances.
Commonly used techniques include Cutout~\cite{devries2017improved} and CutMix~\cite{yun2019cutmix}.
Recent studies have also exploited a limited number of labeled anomalous samples to enhance detection performance, improving upon traditional approaches that rely on simple feature modification. For example, starting from a small set of labeled anomalies, SOEL~\cite{li2023deep} enables the anomaly score ranking learned from labeled data to generalize to unlabeled samples, thereby improving overall detection performance. 
NNG-Mix~\cite{dong2024nng} leverages a small number of labeled samples together with a large amount of unlabeled data to generate synthetic anomalies via nearest-neighbor constraints and Gaussian perturbations.
However, the real anomaly labels required by these methods are not available in the scenario considered in this work. To assess the effectiveness of the proposed approach, we conduct experiments in which SOEL and NNG-Mix are evaluated under pseudo-label settings. The experimental results demonstrate that the proposed PLAG framework remains highly competitive.

\textbf{General Generation Methods.}
General-purpose methods focus on addressing the complex challenges of tabular data, such as missing values, heterogeneity, and data sparsity. 
These methods generate synthetic samples that resemble the original data by modeling the distribution of observed instances~\cite{shi2025comprehensive}. 
Early efforts were primarily based on VAE~\cite{liu2023goggle} and GAN~\cite{xu2019modeling} architectures, which often suffer from issues such as unstable training and mode collapse.
Recently, diffusion models and LLMs have demonstrated strong representation and data understanding capabilities, providing effective solutions to the aforementioned challenges.
For example, TabDDPM~\cite{kotelnikov2023tabddpm} and TabSyn~\cite{zhang2024mixed} adapt the diffusion process to tabular data and to a variational autoencoder–structured latent space, respectively, and both methods outperform traditional GAN- and VAE-based approaches.
GReaT~\cite{borisov2023language}, an LLM-based method, models the distribution of tabular data by conditioning on arbitrary subsets of features and demonstrates strong generative capability.
Although these methods are capable of generating tabular data, they rely on modeling real anomalous samples, which do not align with the label-scarce scenarios of interest in this work. 
More importantly, these approaches are not designed with anomaly detection as their primary objective. They focus on generating synthetic samples that are closest to the reference data, without considering whether such samples effectively enhance the model’s ability to distinguish between normal and anomalous instances.
\begin{figure*}
    \centering
    \includegraphics[width=\textwidth]{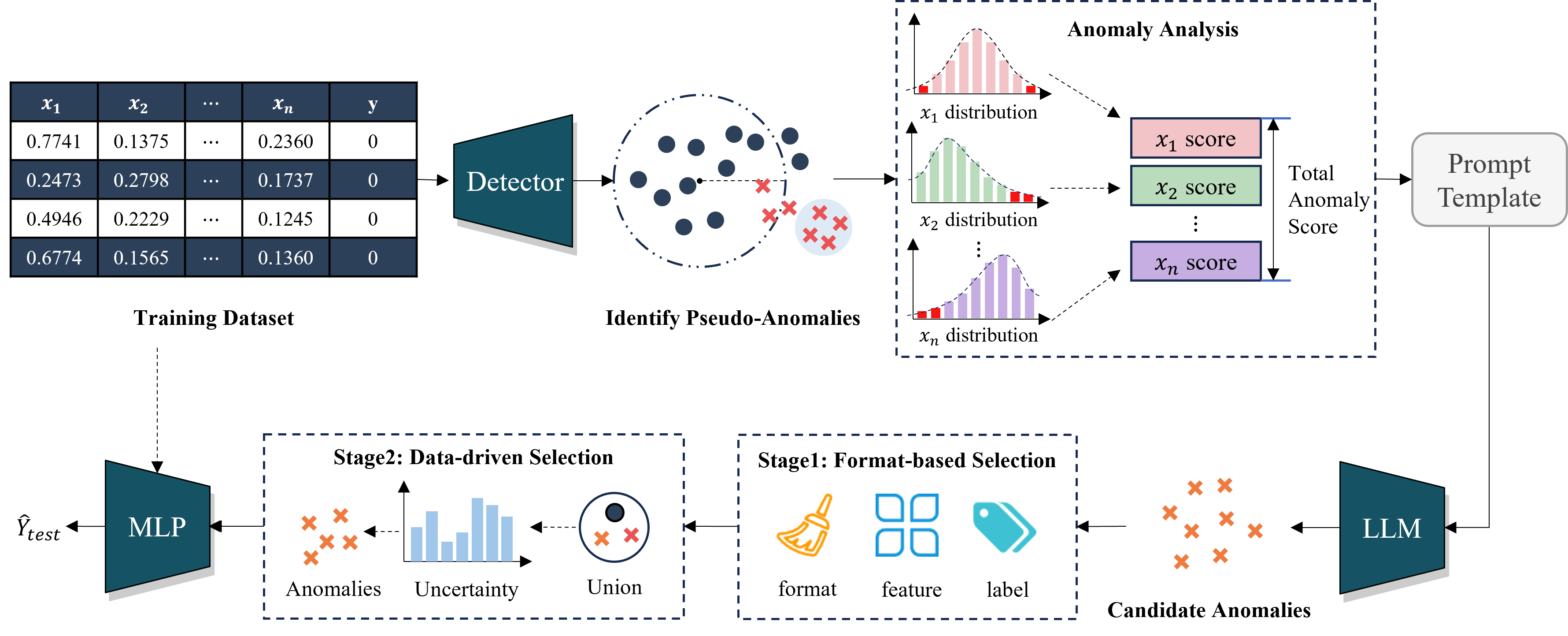}
    \caption{The framework of PLAG.
    (1) Pseudo-Label-Guided Candidate Anomaly Generation.
    PLAG initially employs an unsupervised detector to score unlabeled instances, isolating the top-scoring samples to construct a high-confidence pseudo-anomaly set. 
    Subsequently, we devise a tabular-specific prompt template (as shown in Fig.~\ref{fig:template}). 
    By modeling each feature dimension independently, PLAG decouples global anomaly quantification into an aggregate of feature-level abnormalities. 
    This decoupling mechanism amplifies sensitivity to localized anomalous signals, offering a straightforward yet effective method for directing an LLM to comprehend underlying anomaly semantics and synthesize candidate anomalies.
    (2) Two-Stage Data Selection Strategy. 
    To improve both the fidelity and diversity of the candidate anomalies, PLAG jointly models the normal samples from the training set, the pseudo anomalies and the generated candidate anomalies within a unified information system. 
    By quantifying uncertainty via concept discernibility rather than geometric proximity, it mitigates label ambiguity to robustly select diverse anomalies.
    This design explicitly accounts for the potential diversity of anomalies, and empirical evaluations show that as the dataset size increases, PLAG exhibits superior performance.
    (3) Anomaly detection.
    Utilizing the retained synthetic anomalies alongside the normal samples from the training set, a multilayer perceptron (MLP) model is trained in a supervised manner. The optimized model is subsequently employed to classify whether a given test sample is anomalous.
    }
    \label{fig:framework}
    \vspace{-1em}
\end{figure*}

\section{Methodology}
\subsection{Preliminaries and Problem Definition}
In this work, we investigate how anomaly generation can be leveraged to enhance tabular anomaly detection in scenarios where real anomaly labels are unavailable. Specifically, we consider a training set $\mathcal{D}_{train}=\{\textbf{x}_{i}^{train}\}_{i=1}^{N}$ which consists exclusively of normal samples, and a test set $\mathcal{D}_{test}=\{\textbf{x}_{i}^{test}\}_{i=1}^{N^{\prime}}$ which contains both normal and anomalous samples.
Each sample is associated with a binary label $y\in \{0, 1\}$, where $y=0$ represents a normal sample and $y=1$ represents an anomalous sample.
Here, $N$ and $N'$ indicate the number of samples in $\mathcal{D}_{train}$ and $\mathcal{D}_{test}$, respectively.
The objective of this task is to design an anomaly generation method without requiring manual annotation. 
Specifically, the approach first generates a set of synthetic anomalies, denoted as $\mathcal{D}_{synthetic}$, and subsequently leverages them to augment the training process, thereby enhancing the overall detection performance.

\subsection{The Overview of PLAG}
As illustrated in Fig.~\ref{fig:framework}, PLAG first employs an existing unsupervised anomaly detector to score unlabeled data, assigning pseudo-labels to a subset of instances with high anomaly probabilities. Subsequently, a tabular-specific strategy is devised to decouple the overall anomaly quantification into feature-level abnormalities, effectively characterizing the anomalous semantics. 
Guided by this strategy, tailored prompts direct an LLM to comprehend these patterns and synthesize candidate anomalies.
Nevertheless, since the LLM inevitably generates noisy or redundant samples, directly utilizing these unfiltered instances may compromise the model's discriminative ability.
Therefore, PLAG introduces a two-stage data selection strategy to extract synthetic anomalies that balance fidelity and diversity. 
The first stage conducts format verification to ensure representational validity. 
The second stage then integrates normal samples, pseudo-anomalies, and candidates into a unified information system based on fuzzy rough set theory. 
Here, PLAG quantifies sample uncertainty via concept discernibility rather than mere geometric proximity, mitigating inherent label ambiguity. 
By leveraging these uncertainty metrics, the strategy initially selects synthetic candidates highly dissimilar to normal data to ensure fidelity. 
It then scales the selection threshold toward the normal distribution to incorporate near-boundary hard examples, thereby enhancing generative diversity.
Finally, the filtered synthetic anomalies and normal data are jointly used to train a downstream classifier. 
In our experiments, this classifier is implemented as a simple multilayer perceptron (MLP).

\subsection{Pseudo-Label-Guided Candidate Anomaly Generation}
In scenarios where real labels are unavailable, generating synthetic anomalies without any guidance is unrealistic. 
To address this issue, an existing unsupervised anomaly detection model is employed to learn the normal data distribution from the training set and to infer anomaly probabilities for samples in the test set. 
PLAG then selects samples with high anomaly probabilities and assigns pseudo-anomaly labels to them, which serve as anomaly references. 
This procedure can be formally expressed by Equations~(\ref{eq:existing_model_score}) and~(\ref{eq:pseudo_anomaly}):
\begin{equation}
    \mathcal{S} = \textbf{M}_{E}(\mathcal{D}_{test};\theta^{*}),
    \label{eq:existing_model_score}
\end{equation}
where $M_{E}$ denotes the existing unsupervised anomaly detector, $\mathcal{S}=\{s_{1},s_{2},\cdots,s_{N'}\}$ represents the anomaly probabilities produced by $M_{E}$ on the test set, and $\theta^{*}$ denotes the optimal parameters obtained by training $M_{E}$ on the training dataset.
\begin{equation}
    \mathcal{D}_{pseudo} = \{\textbf{x}_{i} | a_{i}\in Top(\mathcal{S},p_{1}), \textbf{x}_{i}\in \mathcal{D}_{test}\},
    \label{eq:pseudo_anomaly}
\end{equation}
where $\mathcal{D}_{pseudo}$ denotes selected pseudo  anomalies with high anomaly probabilities, $Top$ refers to selecting the top $p_{1}$\% highest scores from the set $\mathcal{S}$.

Leveraging these high-confidence pseudo anomalies, dedicated prompting templates are designed for two-dimensional tabular data to guide a large language model in analyzing anomalous characteristics and generating candidate anomalies. 
Specifically, each instance in a tabular dataset is composed of multiple feature dimensions, and for a given feature dimension, the feature values of all instances form a distribution. 
According to the definition that anomalies are typically dissimilar to the majority of instances in the dataset, feature values that are rare within a distribution are more likely to be anomalous. 
From the perspective of feature distributions, PLAG therefore accumulates the rarity of a sample across different feature dimensions as its anomaly probability. 
Through this explicit analytical formulation, the large language model is guided to understand anomalous patterns and to generate candidate anomalies that exhibit distributional abnormality.

In implementation, the definition of anomalies is first provided to the large language model, which is then guided to analyze the distribution of each feature dimension for given data instances. 
Feature values that appear rarely within the distribution are identified as anomalous, and the anomaly score of the corresponding instance is increased by one.
Here, the identification of rare-value regions is based on the Empirical Cumulative Distribution Function (ECDF). 
ECDF provides a distribution-aware and scale-invariant measure of feature rarity, which enables consistent identification of rare-value regions across heterogeneous features. 
Assuming that each sample is characterized by $m$ feature dimensions, ECDF is defined as shown in Equation (\ref{eq:ecdf}):
\begin{equation}
    \hat{F}_{j}(t) = \frac{1}{m}\sum_{i=1}^{m}\mathbb{I}(\textbf{x}_{ij}<t),
    \label{eq:ecdf}
\end{equation}
where $\hat{F}_{j}(t)$ denotes the proportion of samples whose value on the j-th feature dimension does not exceed $t$, $I(\cdot)$ denotes the indicator function, which equals 1 when the condition inside holds and 0 otherwise.

The rare-value region can then be defined as shown in Equation (\ref{eq:rare_interval}):
\begin{equation}
    \tau_{ij} = 2min\{\hat{F}_{j}(\textbf{x}_{ij}), 1-\hat{F}_{j}(\textbf{x}_{ij})\}
    \label{eq:rare_interval}
\end{equation}
where $\hat{F}_{j}(\mathbf{x}_{ij})$ characterizes the relative position of the sample $\mathbf{x}_{i}$ within the distribution of the j-th feature. $\tau_{ij} \in (0,1]$ denotes the probability that the sample lies in the two-sided tails of the distribution. A smaller value of $\tau_{ij}$ indicates a higher likelihood that the sample falls within a rare-value region.

By combining Equations (\ref{eq:ecdf}) and (\ref{eq:rare_interval}), the anomaly score $s'$ of each sample with respect to the feature distributions can be obtained, as shown in Equation (\ref{eq:new_anomaly_score}):
\begin{equation}
    s'_{i} = \sum_{j=1}^{m}\mathbb{I}(\tau_{ij}<p_{2})
    \label{eq:new_anomaly_score}
\end{equation}
where $s'_{i}$ denotes the resulting anomaly score of sample $i$, and $p_{2}$ is a threshold that determines the degree of rarity of a feature value within the distribution. In the experiments, to avoid manual specification of this parameter, the large language model is allowed to determine it autonomously based on the anomaly definition.

Subsequently, Equations~(\ref{eq:new_pseudo_score}) and (\ref{eq:range}) is used to compute the anomaly scores of the pseudo anomalies and obtain a corresponding score range. 
\begin{equation}
    \mathcal{A}_{pseudo} = \{s'_{i}|\textbf{x}_{i}\in\mathcal{D}_{pesudo}\}
    \label{eq:new_pseudo_score}
\end{equation}
\begin{equation}
    Range = [min(\mathcal{A}_{pseudo}),max(\mathcal{A}_{pseudo})],
    \label{eq:range}
\end{equation}

Finally, as illustrated in Fig.~\ref{fig:template}, we formalize the aforementioned anomaly analysis procedure into a tailored prompt template, denoted as $\mathcal{O}$. 
This template directs the LLM to comprehensively grasp anomalous patterns and synthesize candidate samples that exhibit distinct deviations within the feature distribution. 
Specifically, $\mathcal{O}$ comprises four core components: Task \& Objective, Data Description, Analytical Method, and Output Requirement. A brief description of each is provided below:
\begin{figure}
    \centering
    \includegraphics[width=0.8\columnwidth]{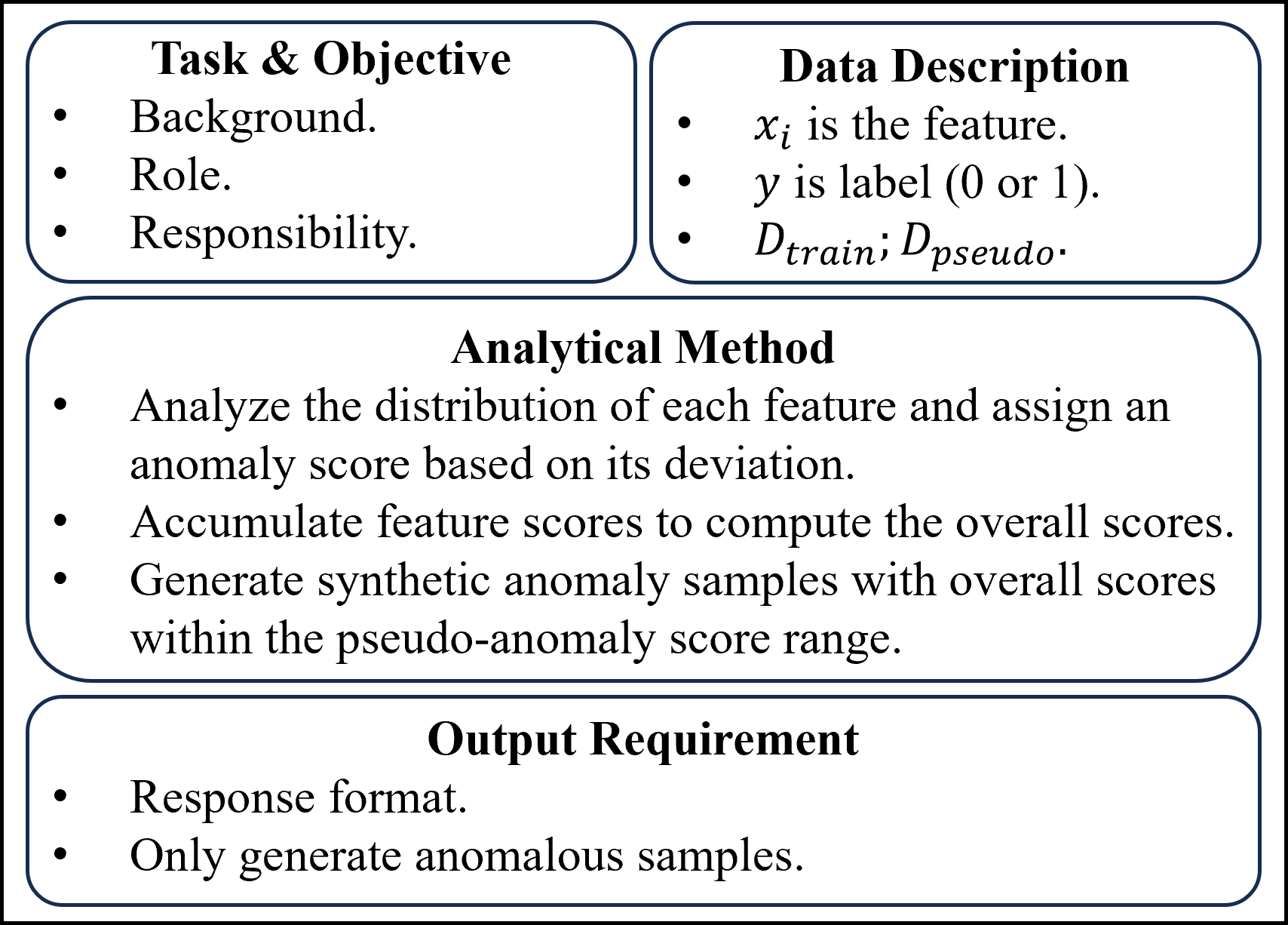}
    \caption{Tabular-specific prompt template for anomaly generation.}
    \label{fig:template}
    \vspace{-1em}
\end{figure}
\begin{itemize}
    \item \textbf{Task \& Objective.} 
    Introduce the background of anomaly detection to the LLM and assign it the role of a data generator, enabling it to understand its responsibilities.

    \item \textbf{Data Description.} 
    Inform the LLM of the composition of the data and provide it with pseudo-anomalies along with a subset of normal samples as references.

    \item \textbf{Analytical Method.} 
    This strategy instructs the LLM to decouple the overall anomaly degree of a sample into an accumulation of feature-level rarities within their respective distributions, thereby guiding it to deeply comprehend the underlying anomalous patterns. 
    Concurrently, by referencing the anomaly score interval of the pseudo-anomalies, it directs the LLM to synthesize candidate samples with closely aligned anomaly scores.
    
    \item \textbf{Output Requirement.} 
    To ensure consistency and eliminate irrelevant content, the LLM is instructed to output only synthetic anomalies in a standardized tabular format, without any explanations or auxiliary text. 
    
\end{itemize}

The generated candidate anomalies are defined in Equation~(\ref{eq:candidate_anomalies}):
\begin{equation}
    \mathcal{D}_{candidate} = LLM(O)
    \label{eq:candidate_anomalies}
\end{equation}

\subsection{Two-Stage Data Selection Strategy}
Although the primary motivation for leveraging pseudo labels to guide anomaly generation is to obtain richer and more diverse synthetic anomalous samples, large language models often exhibit issues such as formatting errors and information redundancy during data generation, which prevent this motivation from being fully realized.
In the experiments, directly incorporating candidate anomalies into model training may not only fail to exploit their data value fully but can even degrade model performance in some cases. 
To address this issue, PLAG proposes a data-driven selection strategy that ensures the fidelity and diversity of the generated data through a two-stage, progressive filtering process.

Specifically, PLAG performs format-based selection in the first stage to examine whether candidate anomalies contain errors in features, labels or other structural aspects. 
In the second stage, PLAG incorporates normal samples, pseudo anomalies and candidate anomalies into a unified information system and evaluates their uncertainty in a data-driven manner.
This design allows the selection process to favor candidate anomalies that are similar to pseudo anomalies while simultaneously ensuring that the selected samples include both instances that are far from the normal distribution and those that are close to it.
This will enhance the diversity of the selected synthetic anomalies.
The format-based selection strategy and the data-driven selection strategy are introduced as follows.

\subsubsection{\textbf{Stage 1: Format-based selection strategy}} 
The strategy validates the format compliance of candidate anomalies by checking whether the output format is correct, the number of features is consistent, the feature values fall within reasonable ranges, and the assigned label appropriately corresponds to an anomaly.
Furthermore, duplicate samples are removed to reduce redundancy in the candidate pool.
The selection process can be formulated as Equation~(\ref{eq:selection_1}).
\begin{equation}
    \mathcal{D}_{selection,1} = Unique(\{x_{i}\in \mathcal{D}_{candidate}|E(x_{i})=1\}),
    \label{eq:selection_1}
\end{equation}
where $Unique$ denotes the operation that removes duplicate samples from the data, and $E$ is the compliance checking function, which is defined in Equation~(\ref{eq:checking_function}).
\begin{equation}
   E(x_{i})=\mathbb{I} (Format(x_{i})\wedge Feature(x_{i})\wedge Label(x_{i})),
    \label{eq:checking_function}
\end{equation}
where $\mathbb{I}$ denotes the indicator function, while Format($\cdot$), Feature($\cdot$), and Label($\cdot$) respectively examine the sample in terms of format consistency, feature validity, and label correctness.

\subsubsection{\textbf{Stage 2: Data-driven selection strategy}}
In this stage, our objective is to ensure the validity of the synthetic anomalies while simultaneously promoting their diversity. 
Specifically, by quantifying sample uncertainty, PLAG initially selects evident anomalies with high uncertainty. 
Subsequently, it relaxes the uncertainty threshold to incorporate additional synthetic anomalies situated closer to the normal distribution.
The crux of achieving this lies in the accurate estimation of sample uncertainty. 
Given the inherent label ambiguity of both pseudo-anomalies and candidate anomalies, relying solely on naive similarity-based methods merely identifies samples that are proximate in the geometric space; it fails to capture the accurate uncertainty of class membership under incomplete or fuzzy knowledge. 

To address this limitation, we introduce the concept of fuzzy rough sets (FRS) to uniformly model the normal samples, pseudo-anomalies, and LLM-generated candidates into a unified information system, as formulated in Equation~(\ref{eq:information_system}). 
Within this system, whether a sample belongs to the anomalous class is no longer determined by a single similarity or distance metric, but is instead characterized by its discernibility with respect to the anomaly concept under the current knowledge structure. 
By distinguishing possible membership from certain membership through upper and lower approximations, fuzzy rough sets enable explicit modeling of the uncertainty between anomalous and normal samples induced by attribute overlap and label ambiguity.
\begin{equation}
    \mathcal{U} = \{\mathcal{D}_{candidate}, \mathcal{D}_{pesudo}, \mathcal{D}_{train}\}.
    \label{eq:information_system}
\end{equation}

Next, we detail how the fuzzy relations are constructed, how information granules are induced, and how uncertainty is quantified to guide the selection of synthetic anomalies.

\textbf{Fuzzy Relation Construction.}
To characterize the discernibility among samples within a unified information system, it is first necessary to define a relation that can reflect the gradual degree of similarity between samples. 
Since normal samples, pseudo-anomalous samples, and candidate anomalies often lack clear boundaries in the attribute space, their class membership is inherently continuous and fuzzy. 
As a result, traditional binary relations or hard-threshold partitioning are inadequate for accurately describing the relational structure among samples. 
Therefore, we introduce a fuzzy relation to quantify the extent to which pairs of samples are similar, which serves as the foundation for subsequent information granule construction and uncertainty modeling.

Given a sample $\textbf{x}_{i}$, Min-Max scaling is first applied to map each feature into the interval $[0,1]$, to remove the effect of scale discrepancies across features, as shown in Equation~(\ref{eq:minmax}):
\begin{equation}
    \mathcal{Z}_{j}(x_{i})=\frac{\mathcal{Z}_{j}(\textbf{x}_{i})-min(\mathcal{Z}_{j})}{max (\mathcal{Z}_{j})-min(\mathcal{Z}_{j})},
    \label{eq:minmax}
\end{equation}
where $\mathcal{Z}_{j}(x_i)$ denotes the value of the $j$-th feature on sample $\textbf{x}_i$, $min(\mathcal{Z}_{j})$ and $max(\mathcal{Z}_{j})$ represent the minimum and maximum values of the $j$-th feature across all samples, respectively.

Subsequently, we employ a Gaussian kernel function to compute pairwise sample similarities and construct the corresponding fuzzy relation. 
By implicitly mapping samples into a high-dimensional feature space, the Gaussian kernel enables the representation of nonlinear structures that are difficult to separate linearly in the original space. 
This kernel-induced similarity measure not only captures local neighborhood relationships among samples but also reflects subtle variations in inter-sample differences on a continuous scale. 
Compared with directly computing distances in a low-dimensional representation space, the Gaussian kernel better preserves the gradual distinctions between anomalous and normal samples in boundary regions, which is particularly important for subsequent uncertainty assessment. 
The specific construction of the fuzzy relation is given in Equation~(\ref{eq:gaussian_kernel}).
\begin{equation}
    r_{ij}=\exp (-\frac{\|\mathcal{Z}(\textbf{x}_i)-\mathcal{Z}(\textbf{x}_j)\|^2}{\delta}),
    \label{eq:gaussian_kernel}
\end{equation}
where $r_{ij}$ denotes the fuzzy relation between sample $\textbf{x}_{i}$ and $\textbf{x}_{j}$, $\delta$ is the Gaussian kernel parameter and  $\|\mathcal{Z}(\textbf{x}_i)-\mathcal{Z}(\textbf{x}_j)\|^{2}$ represents the Euclidean distance between sample $\textbf{x}_{i}$ and $\textbf{x}_{j}$ over all features.

\textbf{Information Granules Induced by Fuzzy Relations.}
After obtaining the fuzzy relations among samples, it is still necessary to further structure these relations to provide operational basic units for uncertainty evaluation. 
Although fuzzy relations can characterize the gradual similarity between arbitrary pairs of samples, they remain at the level of pairwise relationships. 
Thus, they cannot directly reflect the overall semantic position of an individual sample within the entire information system. 
Therefore, we introduce the concept of information granules to organize the dispersed fuzzy relations into structured, sample-centered representations.

As shown in Equation~(\ref{eq:granule}), an information granule characterizes the fuzzy associations between a given sample and all other samples in the information system. By treating the entire set of membership degrees induced by the fuzzy relation as a whole, an information granule jointly considers the relationships between the sample and the entire information system, thereby providing a global, system-level representation.
\begin{equation}
    [x_{i}]=\{r_{i1}, r_{i2}, \cdots, r_{iN_{u}}\},
    \label{eq:granule}
\end{equation}
where $[x_{i}]$ denotes the information granule, $N_{u}$ denotes the number of all samples in $\mathcal{U}$.

The introduction of the information granule $[x_{i}]$ ensures that the sample $x_{i}$ is no longer treated in isolation, but rather as a node embedded in a fuzzy relational network, whose characteristics are jointly determined by its relations with other samples.

\textbf{Upper and Lower approximations.} 
In the information system induced by fuzzy relations, the relationship between samples and the anomaly concept is no longer binary, but instead manifests as a continuous degree of membership. 
Accordingly, to characterize the certain and possible memberships of samples under this fuzzy setting, we adopt the upper and lower approximations from fuzzy rough set theory to model the anomaly concept. 

Specifically, by leveraging existing definitions and theoretical results in the fuzzy rough set literature, we can compute the upper and lower approximations of information granules.

\noindent\textbf{\textit{Definition 1.}} Give a nonempty and finite set $U$, a real-valued function $k: U \times U \rightarrow R$ is said to be a kernel if it is symmetric, that is, $k(x, y)=k(y, x)$ for all $\forall x, y \in U$, and positive-semidefinite.

\noindent\textbf{\textit{Theorem 1.}} Any kernel $k: U \times U \rightarrow[0,1]$ with $k(x, x)=1$ is (at least) $T_{\text {cos }}$-transitive, where
\begin{equation}
    T_{\mathrm{cos}}(a, b)=\max \left(a b-\sqrt{1-a^2} \sqrt{1-b^2}, 0\right)
\end{equation}

As some of the kernel functions are reflexive, $k(x, x)=1$, symmetric $k(x, y)=k(y, x)$, and $T_{\text {cos }}$-transitive, then the relations computed with these kernel functions are fuzzy $T$-equivalence relations.
The Gaussian kernel adopted in our method satisfies these properties.

\noindent\textbf{\textit{Definition 2.}} Given a nonempty universe $U$ and a kernel function $k$ being reflexive, symmetric, and $T_{\text {cos }}$-transitive, for arbitrary fuzzy subset $X \in H(U)$, $H(U)$ denotes the set of all fuzzy relations on $U$, the kernelized fuzzy upper and lower approximations of $X$ with repect to $K_{R}$are defined as follows.
\begin{equation}
    \overline{K_{R}} X(x)=\sup _{y \in U} T_{cos}(k(x, y), X(y))
\end{equation}
\begin{equation}
    \underline{K_{R}} X(x)=\inf _{y \in U} S_{cos}(G(k(x, y)), X(y))
\end{equation}

where
\begin{equation}
    T_{cos}(a, b)=\max \{a b-\sqrt{1-a^2} \sqrt{1-b^2}, 0\},
\end{equation}
\begin{equation}
    S_{cos}(a, b)=\min \{a+b-a b-\sqrt{2 a-a^2} \sqrt{2 b-b^2}, 1\},
\end{equation}
$K_{R}$ denotes the kernelized fuzzy relation, $G(q) = 1- q$ denotes a fuzzy negation operator.

In our work, $K_{R}$ corresponds to $r_{ij}$. 
Therefore, based on \textbf{\textit{\textbf{Definition~2}}} and the information granule $[x_{i}]$ constructed from fuzzy relation $r_{ij}$, we can obtain the lower and upper approximations of $[x_{i}]$, denoted by $\underline{R_{L}}$ and $\overline{R_{U}}$, respectively.
The lower approximation set $\underline{R_{L}}$ contains only those samples in the information system that are highly similar to the sample $x_i$. 
In contrast, the upper approximation set $\overline{R_{U}}$ relaxes the similarity threshold and therefore includes some additional samples that exhibit slightly lower similarity. 

\textbf{Uncertainty Quantification.} 
We then compute the ratio between the lower approximation and the upper approximation to obtain the approximation accuracy, as shown in Equation~(\ref{eq:uncertainty_1}):
\begin{equation}
    \alpha([x_i])=\frac{|\underline{R_{L}}[x_i]|}{|\overline{R_{U}}[x_i]|} 
    \label{eq:uncertainty_1}
\end{equation}

The approximation accuracy reflects the proportion of certainty within possibility, and for a given sample 
$[x_{i}]$, it characterizes the uncertainty associated with that sample. 
A higher value of $ \alpha([x_i])$ indicates greater certainty of sample $x_{i}$, whereas a lower value implies higher uncertainty.
However, we observe that Equation~~(\ref{eq:uncertainty_1}) only accounts for the uncertainty of sample $x_{i}$ within its corresponding information granule $[x_{i}]$, while neglecting the proportion of the class to which $x_{i}$ belongs in the entire information system. 
This aspect is particularly important in anomaly detection, since anomalies are typically rare in the dataset.
If two samples with identical approximation accuracy $\alpha$ are treated equally without considering the proportion of their corresponding classes in the information system, it becomes difficult to distinguish whether these samples are normal or anomalous.
For example, consider an anomalous sample $\textbf{x}_{1}$ with $|\underline{R_{L}}|=1$ and $|\overline{R_{U}}=2$, yielding an approximation accuracy of $\alpha=0.5$.
In contrast, a normal sample $\textbf{x}_{2}$ with  $|\underline{R_{L}}|=50$ and $|\overline{R_{U}}=100$, also yielding an approximation accuracy of $\alpha=0.5$.
In this case, $\alpha$ fails to reflect the underlying differences in sample rarity, thereby limiting its effectiveness in anomaly detection.

To address this issue, PLAG introduces a weighting factor $\lambda$. 
It adjusts the uncertainty score of sample $x$ by incorporating the ratio between the magnitude of its information granule and the total number of samples in the system. The revised computation is shown in Equation~(\ref{eq:uncertainty_2}):
\begin{equation}
    \alpha^{\prime}([x_i])=1-\lambda\cdot \frac{|\underline{R_{L}}[x_i]|}{|\overline{R_{U}}[x_i]|}, \lambda = \frac{|[x_i]|}{|\mathcal{U}|}.
    \label{eq:uncertainty_2}
\end{equation}

In the revised formulation, a larger value of $\alpha^{\prime}$ indicates higher uncertainty of the sample, suggesting a greater likelihood of being anomalous.
After optimization, the score of sample $\textbf{x}_{1}$ approaches 1, while the score of sample $\textbf{x}_{2}$ approaches 0.5, effectively distinguishing these two samples.

\textbf{Uncertainty-Guided Data Selection.}
Selecting only anomalies that are far from the normal distribution can ensure the effectiveness of the synthesized data, but it tends to result in a limited diversity of anomaly patterns, causing the model to overemphasize extreme anomalies while neglecting more challenging boundary anomalies commonly encountered in real-world scenarios. 
In contrast, moderately incorporating anomalous samples that lie close to the normal distribution helps characterize the fuzzy boundary between normality and abnormality, thereby enhancing the model’s generalization ability to complex anomaly patterns. 
Therefore, when selecting synthetic anomalous samples, PLAG aims not only to introduce evident anomalies that significantly deviate from the normal distribution, but also to retain a subset of synthesized samples that are close to the normal distribution yet exhibit potential anomalous behavior under the current knowledge structure.

To achieve this goal, PLAG first exploits the $\alpha'$ values of all samples in the set $\mathcal{U}$ obtained in the preceding analysis and applies the $3\sigma$ rule to establish a strict initial threshold, thereby selecting synthetic samples that are far from the normal distribution to ensure the effectiveness of the selected anomalies. 
Subsequently, by referring to the $\alpha'$ value range of pseudo-anomalous samples, the selection threshold is gradually relaxed, allowing the selected anomalies to progressively approach the normal distribution and thus increasing the diversity of the synthesized anomalies.
\begin{equation}
    \mathcal{D}_{selected,2} = \{\textbf{x}_{i} | \alpha^{\prime}([x_i]_)\ge p_{3}, \textbf{x}_{i}\in \mathcal{D}_{selection,1}\},
    \label{eq:selection_2}
\end{equation}

In the equation, $p_{3}$ serves as a threshold to control the diversity of selected anomalies. It's value is initially set to $mean+3\sigma$, where $mean$ denotes the average of the values of $\alpha'$ on all samples in $\mathcal{U}$, and $\sigma$ represents the corresponding standard deviation. 
Subsequently, the value of $p_{3}$ is adjusted to select a more diverse set of anomalies, with its specific value determined by the number of samples selected in the experiments.

\subsection{Anomaly Detection} \label{method:detection}
Finally, the selected synthetic anomalies $\mathcal{D}_{selection,2}$ and the training set $\mathcal{D}_{train}$ are used to train an model with the Focal Loss~\cite{lin2017focal}, as defined in Equations~(\ref{eq:discriminator}) and~(\ref{eq:loss}).
In our experiments, we employ a simple multilayer perceptron as a classifier.
\begin{equation}
    \hat{Y}_{test} = MLP(\mathcal{D}_{train},\mathcal{D}_{selection,2}; \theta^{*})
    \label{eq:discriminator}
\end{equation}
\begin{equation}
\begin{aligned}
     L(\theta) = - \frac{1}{N_{u}} \sum_{i=1}^{N_{u}} ( \alpha (1 - \hat{y}_i)^\gamma y_i \log(\hat{y}_i) \\
     + (1 - \alpha) \hat{y}_i^\gamma (1 - y_i) \log(1 - \hat{y}_i) )
    \label{eq:loss}
\end{aligned}
\end{equation}
where $\theta^{*}$ is the optimal parameter obtained in training, $y_{i}$ denotes the true label of sample $x_{i}^{test}$, the true labels of all synthetic anomalies are set to 1, and $\hat{y}_{i}$ is the predicted label of sample $x_{i}^{test}$, $\alpha$ and $\gamma$ are the parameters of Focal Loss.

\section{Experiments}
\subsection{Experimental Setup}
\subsubsection{Datasets.}
To rigorously evaluate the anomaly detection performance and robustness of the proposed model, we construct a comprehensive evaluation suite drawing from the authoritative ADBench~\cite{han2022adbench}. 
Given the stringent input and output context window limitations inherent to Large Language Models (LLMs), exhaustive evaluation across all available datasets is computationally prohibitive. Therefore, instead of employing homogeneous data, we purposefully select a representative subset of datasets presenting diverse levels of difficulty.
These datasets span across distinct extremes in data distribution, with sample sizes ranging from 278 to 3772, and anomaly ratios exhibiting severe class imbalance (from 2.5\% to 35\%). 
Furthermore, these datasets originate from highly complex real-world domains, specifically healthcare and biology, which are notoriously characterized by high feature heterogeneity and sophisticated underlying patterns, thereby serving as an ideal testbed for cross-scenario generalization.
The statistics of these datasets are shown in TABLE~\ref{tab:datasets}. 
\begin{table}[h]
    \centering
        \caption{The statistics of datasets.}
    \begin{tabular}{cccc}
    \hline
    \textbf{Dataset} & \textbf{Samples} & \textbf{Dim} & \textbf{Anomaly} \\ 
    \hline
       Wbc & 278 & 30 & 21 (5.6\%) \\
       Pima & 768 & 8 & 268 (35\%) \\
       Cardio & 1831 & 21 & 176 (9.6\%) \\
       Yeast & 1484 & 8 & 527 (34\%) \\
       Thyroid & 3772 & 6 & 93 (2.5\%) \\
    \hline
    \end{tabular}
    \label{tab:datasets}
\end{table}

\subsubsection{Baselines.} 
To comprehensively evaluate the performance of the proposed approach, we compare it against eight baseline methods. 
These baselines are strategically categorized into three groups, covering a wide spectrum from traditional heuristic data augmentation to state-of-the-art deep generative models: 
(1) three widely used heuristic data augmentation techniques (Cutout~\cite{devries2017improved}, CutMix~\cite{yun2019cutmix}, and Gaussian noise); 
(2) two advanced, label-dependent anomaly generation methods (NNG-Mix~\cite{dong2024nng} and SOEL~\cite{li2023deep}); 
and (3) three state-of-the-art general-purpose tabular data generation approaches (TabDDPM~\cite{kotelnikov2023tabddpm}, GReaT~\cite{borisov2023language}, and TabSyn~\cite{zhang2024mixed}), which represent cutting-edge diffusion and LLM-based techniques.

Notably, NNG-Mix and SOEL inherently rely on a limited number of ground-truth anomalous samples to guide their generation process. 
Similarly, general-purpose synthesizers such as TabDDPM, GReaT, and TabSyn require labeled training data to effectively model specific conditional distributions. 
Consequently, without the reference of genuine anomalies, these baselines fail to spontaneously synthesize anomalous instances. 
To reconcile this within our unsupervised setting, we replace all requisite ground-truth anomaly labels with pseudo-labels generated by existing unsupervised detectors. 
This adaptation perfectly aligns with the operational conditions of our proposed framework, thereby ensuring a rigorously fair comparison in realistic scenarios where ground-truth labels are unavailable.
A brief description of each method is provided as follows:
\begin{itemize}
    \item \textbf{Cutout:} Cutout generates new samples by randomly masking square or contiguous regions within a sample.

    \item \textbf{CutMix:} CutMix generates new samples by cutting and pasting patches among training samples, while proportionally mixing the corresponding ground-truth labels according to the patch areas.

    \item \textbf{Gaussian noise:} Gaussian noise constructs new anomalous samples by injecting randomly sampled Gaussian noise into reference anomalous instances.

    \item \textbf{NNG-Mix:} NNG-Mix leverages a small number of labeled samples, combined with a large amount of unlabeled data, to generate synthetic anomalies through nearest neighbor constraints and Gaussian perturbations. 

    \item \textbf{SOEL:} SOEL leverages a small set of labeled anomalies to learn a ranking-based anomaly scoring function that generalizes to unlabeled samples.

    \item \textbf{TabDDPM:} TabDDPM generates high-quality synthetic tabular samples by modeling numerical and categorical features through separate diffusion denoising processes. 

    \item \textbf{GReaT:} GReaT converts tabular data into natural language sequences that encode feature semantics and then employs an autoregressive language model to generate synthetic samples.

    \item \textbf{TabSyn:} TabSyn embeds mixed-type tabular data into a continuous latent space using a VAE and generates samples by applying a score-based diffusion model in the latent space.
    
\end{itemize}

\begin{table*}[t]
\centering
\caption{Quantitative evaluation of anomaly detection performance across all baseline data generation methods. The best and second-best results are denoted in bold and underlined, respectively.}
\setlength{\tabcolsep}{1mm}
\begin{tabular}{c|ccc|ccc|ccc|ccc|ccc}
\hline
\multirow{2}{*}{\textbf{Method/Dataset}} & \multicolumn{3}{c|}{\textbf{Wbc}} & \multicolumn{3}{c|}{\textbf{Pima}} & \multicolumn{3}{c|}{\textbf{Cardio}} & \multicolumn{3}{c}{\textbf{Yeast}} & \multicolumn{3}{c}{\textbf{Thyroid}}\\ 
\cline{2-16} 
 & \textbf{auc-roc} & \textbf{auc-pr} & \textbf{f1}
 & \textbf{auc-roc} & \textbf{auc-pr} & \textbf{f1}
 & \textbf{auc-roc} & \textbf{auc-pr} & \textbf{f1}
 & \textbf{auc-roc} & \textbf{auc-pr} & \textbf{f1}
 & \textbf{auc-roc} & \textbf{auc-pr} & \textbf{f1}\\
 \hline

\textbf{Cutout} 
& 0.4979 & 0.3338 & 0.3309
& 0.4691 & 0.5299 & 0.6802
& 0.5532 & 0.3134 & 0.3711
& 0.5282 & 0.5207 & 0.6768
& 0.9710 & 0.7595 & 0.7517 \\

\textbf{CutMix} 
& 0.4748 & 0.3154 & 0.3477
& 0.4974 & 0.5483 & 0.6863
& 0.7041 & 0.4206 & 0.4439
& 0.5306 & 0.5242 & 0.6772
& 0.4706 & 0.1776 & 0.2199 \\

\textbf{Gaussian noise} 
& 0.9639 & 0.8224 & 0.7787
& 0.5176 & 0.5574 & \underline{0.6886}
& 0.7515 & 0.4658 & 0.4755
& 0.4928 & 0.4946 & 0.6766  
& 0.9491 & 0.5975 & 0.6430 \\

\textbf{NNG-Mix} 
& 0.8975 & 0.4689 & 0.3871
& \textbf{0.7153} & 0.5594 & 0.4393
& 0.7696 & 0.5410 & 0.5591
& 0.5458 & 0.3796 & 0.3676
& \textbf{0.9943} & 0.7613 & 0.5185 \\

\textbf{SOEL} 
& 0.9499 & 0.8338 & 0.7142
& 0.5679 & 0.6134 & 0.5447
& 0.7705 & 0.6512 & 0.6136
& 0.5195 & 0.5244 & 0.5259
& 0.8205 & 0.3411 & 0.3720 \\

\textbf{TabDDPM} 
& 0.9402 & 0.8791 & 0.7906
& 0.5687 & 0.6159 & 0.6870
& 0.9668 & \underline{0.8807} & 0.7877
& 0.4483 & 0.4658 & 0.6750
& 0.9705 & 0.7667 & 0.7367 \\

\textbf{GReaT} 
& 0.9080 & 0.8133 & 0.7499
& 0.5314 & 0.5476 & 0.6812
& 0.8609 & 0.7048 & 0.6681
& 0.5538 & 0.5250 & 0.6771
& 0.9723 & 0.7369 & 0.7253 \\

\textbf{TabSyn} 
& 0.9463 & 0.8939 & \underline{0.8000}
& 0.5628 & 0.6124 & 0.6833
& 0.9327 & 0.7853 & 0.6860
& 0.4885 & 0.4804 & 0.6756 
& 0.9750 & 0.7968 & 0.7660 \\

\cdashline{1-16}

\textbf{W/ GPT-3.5-turbo} 
& \underline{0.9710} & 0.8649 & 0.7550
& \underline{0.6596} & \underline{0.6838} & \textbf{0.7023}
& \underline{0.9695} & 0.8747 & 0.7964
& \textbf{0.5982} & \textbf{0.5711} & \textbf{0.6807}
& \underline{0.9796} & \underline{0.8464} & \underline{0.7986} \\

\textbf{w/ GPT-4} 
& 0.9622 & \underline{0.8817} & 0.7826
& 0.6251 & 0.6782 & 0.6876
& 0.9681 & \textbf{0.8884} & 0.7927 
& \underline{0.5596} & \underline{0.5433} & \underline{0.6774}
& \underline{0.9796} & \textbf{0.8680} & \textbf{0.8241} \\

\textbf{w/ DeepSeek-V2.5} 
& 0.9669 & 0.8639 & 0.7790 
& 0.5937 & 0.6440 & 0.6846
& 0.9675 & 0.8647 & \underline{0.7995} 
& 0.5117 & 0.5088 & 0.6769  
& 0.9742 & 0.8256 & 0.7838 \\

\textbf{w/ Qwen3-14B} 
& \textbf{0.9779} & \textbf{0.9019} & \textbf{0.8275} 
& 0.6271 & \textbf{0.6952} & 0.6828
& \textbf{0.9715} & 0.8677 & \textbf{0.8170} 
& 0.5259 & 0.5253 & 0.6759  
& 0.8762 & 0.3279 & 0.4449 \\

\hline
\end{tabular}
\label{tab:generation}
\end{table*}

\subsubsection{Metrics.}
Following prior studies~\cite{yin2024mcm,ye2025drl}, we employ three comprehensive metrics (AUC-ROC, AUC-PR, and F1 score) to evaluate the anomaly detection performance of all methods.

\subsubsection{Implementation Details.}
All experiments are conducted on NVIDIA GeForce RTX 2080Ti (12GB) with PyTorch~\cite{paszke2019pytorch}, and the version of the operating system is Ubuntu 16.04.
The unsupervised anomaly detection method, MCM, is employed as the default detector to provide the initial pseudo labels.
In the main results, PLAG utilizes various large language models to assess their impact on anomaly generation. 
However, considering the balance between performance and usage cost, GPT-3.5 is adopted as the default large language model in experiments beyond the main evaluation.
For each tabular dataset, GPT-3.5 is used to generate 100 synthetic anomalies. When the number of samples in a dataset exceeds 3,000, an additional 100 anomalies are generated. An equal number of synthetic anomalies is used across all methods to ensure a fair experimental comparison.
To strictly validate the efficacy of the synthesized anomalies, a vanilla multilayer perceptron is adopted as the downstream supervised classifier, trained using the synthesized anomalies alongside original normal samples from the training set. 
The deliberate choice of such a straightforward network architecture aims to preclude performance gains derived from complex classifier capacities, thereby directly demonstrating the superior quality of our generation framework. 
The model is optimized using the Focal Loss, with the number of training epochs set to 250, a batch size of 256, and a learning rate of 0.001.
The results reported in the main experimental section are obtained by averaging over five independent runs, whereas the results of the other experiments are derived from a single independent run.

\subsection{Main Results}
We comprehensively evaluate the tabular anomaly detection performance of our PLAG framework across different LLM integrations, benchmarking it against eight baseline methods. 
As extensively detailed in TABLE~\ref{tab:generation}, the experimental results demonstrate the consistent superiority of our approach across diverse datasets.

A closer inspection of the results reveals that while NNG-Mix attains the best AUC-ROC scores among the baselines on the Pima and Thyroid datasets, its performance drops precipitously in terms of AUC-PR and F1 score. 
This stark contrast reveals that NNG-Mix's high AUC-ROC is largely driven by a general tendency to assign higher global ranking scores to anomalies, rather than demonstrating precise identification of the minority class. 
Crucially, this observation corroborates the consensus that in imbalanced tabular scenarios, an inflated AUC-ROC can be highly deceptive and does not inherently signify a superior anomaly detection capability.
In contrast, although our PLAG framework yields a marginally lower AUC-ROC than NNG-Mix on these two datasets, it demonstrates superiority on the more discriminative metrics of AUC-PR and F1 score. 
Specifically, configuring GPT-3.5-turbo as the default generative model, PLAG outperforms NNG-Mix on the Pima dataset by substantial margins of 0.1244 and 0.2630 in AUC-PR and F1 score, respectively. 
Similar remarkable improvements of 0.0851 and 0.2801 are observed on the Thyroid dataset. 
These results solidly demonstrate that the synthetic anomalies generated by PLAG are more effective at assisting the downstream model in unearthing and accurately capturing latent anomalous instances within the test set.

It is worth emphasizing that PLAG variants equipped with different LLMs consistently achieve state-of-the-art AUC-PR and F1 scores across all datasets. 
This fully demonstrates that PLAG can provide highly effective support for tabular anomaly detection tasks by generating precise and high-quality synthetic anomalies. 
Meanwhile, we observe an interesting phenomenon in our experiments: the parameter scale and version iteration of the LLMs do not exhibit a strictly positive correlation with the final anomaly detection performance. 
This is likely because the synthesis of tabular anomalies relies more heavily on the model's strict adherence to local numerical boundaries and underlying structural distributions, rather than the complex natural language reasoning capabilities inherent to larger or newer models. 
This also indirectly illustrates that PLAG is not highly dependent on the most cutting-edge LLMs; instead, by virtue of its effective generation guidance strategies and quality constraint mechanisms, it can acquire synthetic data that better aligns with the anomaly detection task.
\begin{figure}[t]
    \centering
    \includegraphics[width=\columnwidth]{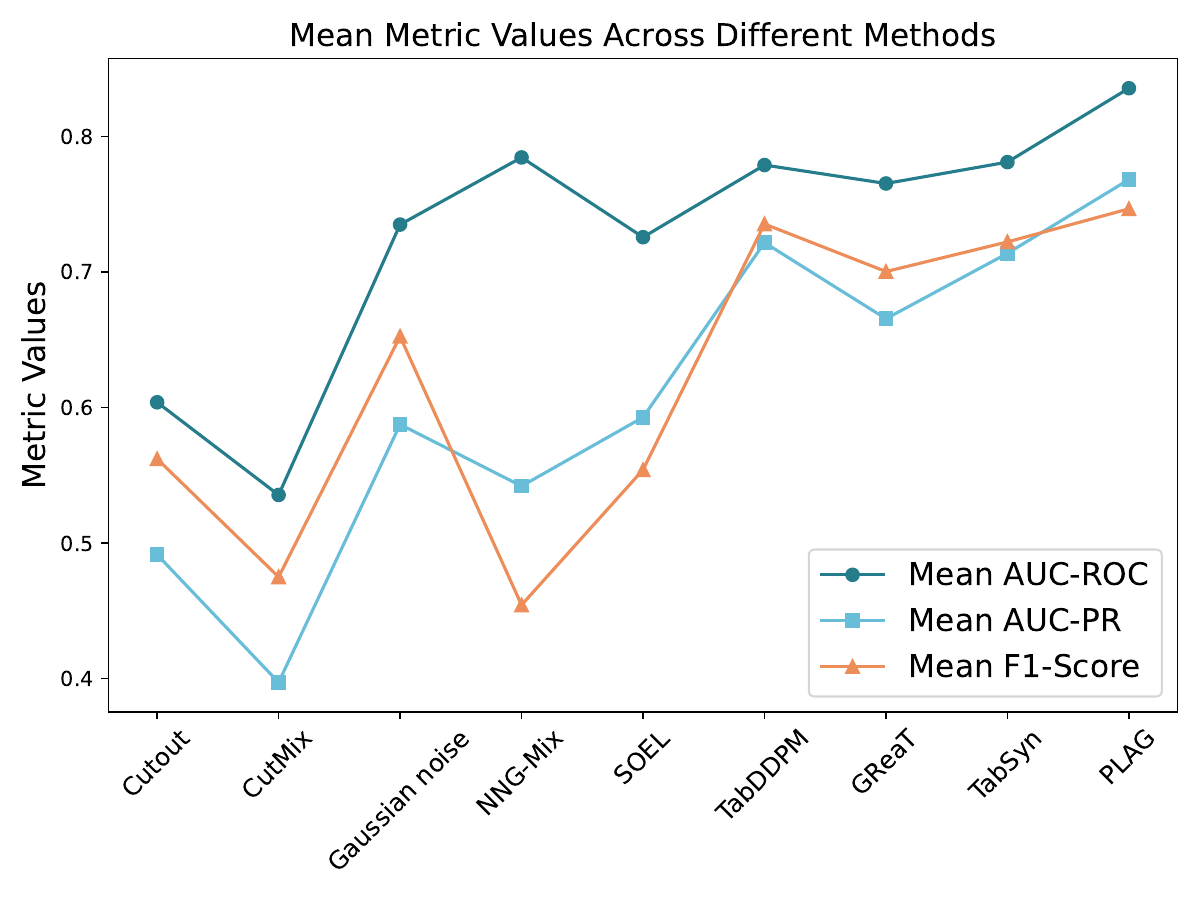}
    \caption{Average performance metrics of various methods across all datasets.}
    \label{fig:mean_metric}
    \vspace{-1em}
\end{figure}

Furthermore, we plot the average metrics of each method across all datasets, as shown in Fig.~\ref{fig:mean_metric}, configuring PLAG with GPT-3.5-turbo to strike a balance between cost and efficacy. 
The results indicate that the three heuristic augmentation techniques exhibit the poorest overall performance. 
Interestingly, Gaussian noise injection unexpectedly outperforms both Cutout and CutMix. 
This is likely because noise-perturbed pseudo-anomalies retain their original anomalous representations, whereas the direct feature modifications inherent in Cutout and CutMix inadvertently dilute the anomaly severity, generating less anomalous instances.
The two label-dependent anomaly generation methods (NNG-Mix and SOEL) achieve excellent AUC-ROC values, but their AUC-PR and F1 scores are remarkably low. 
This indicates that while these approaches are capable of ranking the anomaly scores of most anomalous samples ahead of normal ones globally, they fall short in their capacity to accurately identify the actual anomalies.
The three general-purpose synthesizers outperform the previous two categories of methods. 
This indicates that although they are not explicitly designed for anomaly generation, they possess a robust capability to model tabular data distributions. 
If sufficient ground-truth anomalies were available as references, these methods could serve as highly competitive options. 
Unfortunately, the inherent scarcity of anomalous samples and the notorious difficulty in obtaining true labels in anomaly detection tasks severely restrict their potential.

Compared to the aforementioned baselines, our proposed PLAG achieves optimal overall performance across all three evaluation metrics. 
This substantial enhancement is primarily attributed to our carefully designed dual-mechanism approach. 

\subsection{Compatibility and Enhancement with Diverse Base Detectors}
\begin{table*}[t]
\centering
\caption{The evaluation results of different methods before and after integration into PLAG. Here, "model$^{+}$" indicates that the corresponding model has been incorporated into PLAG.}
\setlength{\tabcolsep}{1mm}
\begin{tabular}{c|ccc|ccc|ccc|ccc|ccc}
\hline
\multirow{2}{*}{\textbf{Method/Dataset}} & \multicolumn{3}{c|}{\textbf{Wbc}} & \multicolumn{3}{c|}{\textbf{Pima}} & \multicolumn{3}{c|}{\textbf{Cardio}} & \multicolumn{3}{c}{\textbf{Yeast}} & \multicolumn{3}{c}{\textbf{Thyroid}}\\ 
\cline{2-16} 
 & \textbf{auc-roc} & \textbf{auc-pr} & \textbf{f1}
 & \textbf{auc-roc} & \textbf{auc-pr} & \textbf{f1}
 & \textbf{auc-roc} & \textbf{auc-pr} & \textbf{f1}
 & \textbf{auc-roc} & \textbf{auc-pr} & \textbf{f1}
 & \textbf{auc-roc} & \textbf{auc-pr} & \textbf{f1}\\
\hline

\textbf{OCSVM} 
& 0.8269 & 0.7334 & 0.7143 
& 0.5000 & 0.7587 & 0.6819
& 0.8915 & 0.7765 & 0.7476 
& 0.4691 & 0.5624 & 0.3697 
& 0.5891 & 0.2713 & 0.2420 \\

\textbf{OCSVM$^{+}$} 
& 0.9731 & 0.9020 & \textbf{0.8148} 
& 0.6291 & 0.6492 & \textbf{0.6862} 
& 0.9723 & 0.8752 & \textbf{0.8243} 
& 0.5891 & 0.5614 & \textbf{0.6759} 
& 0.5000 & 0.7404 & \textbf{0.6495} \\

\cdashline{1-16}

\textbf{DeepSVDD} 
& 0.7957 & 0.5747 & 0.4394
& 0.5579 & 0.6650 & 0.5603 
& 0.6117 & 0.4439 & 0.4317 
& 0.4753 & 0.5660 & 0.3708 
& 0.7598 & 0.5543 & 0.5853 \\

\textbf{DeepSVDD$^{+}$} 
& 0.9662 & 0.8895 & \textbf{0.7921} 
& 0.6530 & 0.6655 & \textbf{0.6865} 
& 0.9114 & 0.7448 & \textbf{0.6858} 
& 0.5445 & 0.5304 & \textbf{0.6765} 
& 0.9301 & 0.6214 & \textbf{0.6035} \\

\cdashline{1-16}

\textbf{GOAD} 
& 0.9462 & 0.7727 & 0.7121
& 0.6423 & 0.6525 & 0.6194
& 0.6617 & 0.4949 & 0.4256
& 0.6019 & 0.5884 & 0.5964 
& 0.9169 & 0.4624 & 0.4468 \\

\textbf{GOAD$^{+}$} 
& 0.9681 & 0.8906 & \textbf{0.8049} 
& 0.6134 & 0.6327 & \textbf{0.6928} 
& 0.9205 & 0.7933 & \textbf{0.7217} 
& 0.5598 & 0.5573 & \textbf{0.6781} 
& 0.9815 & 0.8382 & \textbf{0.8086} \\

\cdashline{1-16}

\textbf{NeuTral AD} 
& 0.7927 & 0.3452 & 0.3939 
& 0.5592 & 0.5604 & 0.5746 
& 0.7331 & 0.3988 & 0.4463 
& 0.5567 & 0.5488 & 0.5610
& 0.9719 & 0.8099 & 0.7482 \\

\textbf{NeuTral AD$^{+}$} 
& 0.9682 & 0.8632 & \textbf{0.7679} 
& 0.6542 & 0.6824 & \textbf{0.6850} 
& 0.9578 & 0.8487 & \textbf{0.7778} 
& 0.5347 & 0.5117 & \textbf{0.6777} 
& 0.8755 & 0.7684 & \textbf{0.7695} \\

\cdashline{1-16}

\textbf{ICL} 
& 0.9301 & 0.7925 & 0.6818 
& 0.6152 & 0.6152 & 0.6045 
& 0.8144 & 0.6792 & 0.5593 
& 0.4940 & 0.4975 & 0.4980 
& 0.8360 & 0.8659 & 0.7482 \\

\textbf{ICL$^{+}$} 
& 0.9719 & 0.8844 & \textbf{0.7947}
& 0.6588 & 0.6717 & \textbf{0.6870}
& 0.9720 & 0.8804 & \textbf{0.8104} 
& 0.5465 & 0.5359 & \textbf{0.6791} 
& 0.9759 & 0.8209 & \textbf{0.7901} \\

\cdashline{1-16}

\textbf{DIF} 
& 0.9434 & 0.7545 & 0.6666 
& 0.6045 & 0.6330 & 0.5982 
& 0.9592 & 0.8356 & 0.7759 
& 0.4112 & 0.4659 & 0.4370  
& 0.9375 & 0.5510 & 0.5496\\

\textbf{DIF$^{+}$} 
& 0.9857 & 0.9131 & \textbf{0.8184} 
& 0.6499 & 0.6745 & \textbf{0.6967} 
& 0.9800 & 0.9160 & \textbf{0.8408} 
& 0.4849 & 0.4984 & \textbf{0.6796} 
& 0.9795 & 0.8112 & \textbf{0.8008} \\

\cdashline{1-16}

\textbf{SLAD} 
& 0.9007 & 0.6176 & 0.5606 
& 0.5615 & 0.5861 & 0.5646
& 0.8357 & 0.6781 & 0.5988 
& 0.5103 & 0.5163 & 0.5072  
& 0.9673 & 0.7904 & 0.7127\\

\textbf{SLAD$^{+}$} 
& 0.9795 & 0.8885 & \textbf{0.7896} 
& 0.5803 & 0.6166 & \textbf{0.6811} 
& 0.8841 & 0.7383 & \textbf{0.6856} 
& 0.5542 & 0.5471 & \textbf{0.6761} 
& 0.9807 & 0.8393 & \textbf{0.8014} \\

\cdashline{1-16}
 
\textbf{MCM} 
& 0.9814 & 0.8854 & 0.7329
& 0.6884 & 0.6859 & 0.6418
& 0.9360 & 0.8489 & 0.6818
& 0.4571 & 0.4881 & 0.4862
& 0.9821 & 0.8306 & 0.7624 \\

\textbf{MCM$^{+}$} 
& 0.9710 & 0.8649 & \textbf{0.7550}
& 0.6596 & 0.6838 & \textbf{0.7023}
& 0.9695 & 0.8747 & \textbf{0.7964}
& 0.5982 & 0.5711 & \textbf{0.6807}
& 0.9796 & 0.8464 & \textbf{0.7986} \\

\hline
\end{tabular}
\label{tab:adaptability}
\end{table*}
A robust anomaly generation framework should not be overly contingent upon a single, specific pseudo-labeling algorithm. 
To validate the model-agnostic flexibility and broad applicability of our PLAG framework, we investigate its compatibility with a diverse array of off-the-shelf unsupervised anomaly detection methods in this section. 
Specifically, we purposefully select several representative existing models, spanning different technical paradigms, to serve as the initial pseudo-label providers (i.e., base detectors). 
Through this diversified evaluation setup, our primary motivation is to verify whether PLAG can consistently act as a plug-and-play "performance amplifier".

To this end, we integrate PLAG with eight representative unsupervised anomaly detection methods and systematically evaluate their performance dynamics before and after incorporating our framework. 
The comprehensive results, summarized in Table~\ref{tab:adaptability}, compellingly validate our hypothesis. 
As clearly illustrated, all eight methods exhibit substantial performance leaps when augmented by PLAG. 
These consistent improvements demonstrate that by synthesizing high-quality and diverse anomalies, PLAG successfully mitigates the performance bottlenecks conventionally imposed by the absence of true anomaly labels. 
Furthermore, the universally positive gains across distinct algorithmic paradigms firmly establish PLAG as a highly flexible and generalizable enhancement framework, capable of being seamlessly integrated with existing tabular anomaly detection pipelines.
\begin{figure}[t]
    \centering
    \includegraphics[width=\columnwidth]{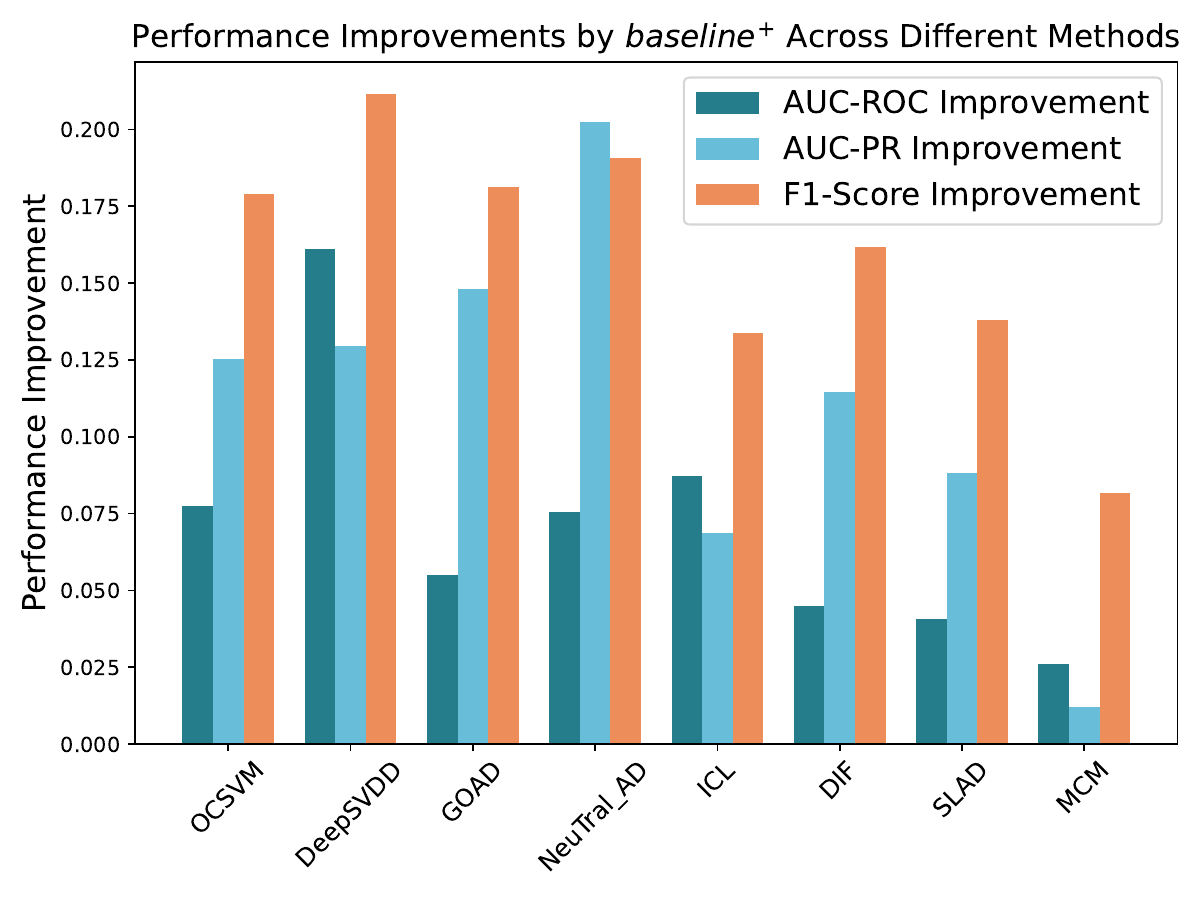}
    \caption{The performance gains introduced by PLAG across the diverse baselines.}
    \label{fig:mean_metric_improvement}
    \vspace{-1em}
\end{figure}

Furthermore, to provide a more intuitive illustration of its enhancement capabilities, we visualize the performance gains introduced by PLAG across the diverse baselines in Fig.~\ref{fig:mean_metric_improvement}. 
The quantitative results show that the integration of PLAG yields remarkable improvements ranging from 0.0261 to 0.1610 in AUC-ROC, 0.0121 to 0.2023 in AUC-PR, and 0.0818 to 0.2114 in the F1-score. 
These quantitative leaps corroborate that PLAG operates as a effective, generalized catalyst for tabular anomaly detection, capable of injecting substantial and consistent performance gains into a wide spectrum of existing detectors through robust anomaly synthesis.

\subsection{Ablation Study}
To validate the internal mechanisms of our framework, we conduct an ablation study by constructing two variants of PLAG. 
This aims to independently evaluate the contributions of its two core components: Pseudo-Label-Guided Candidate Anomaly Generation and the Two-Stage Data Selection Strategy. 
Specifically, the variant denoted as \textit{w/o generation} entirely bypasses the generative module, training the downstream classifier solely on the initial pseudo-labeled anomalies and original training data. 
Conversely, the variant \textit{w/o selection} omits the filtering mechanism, training the classifier directly on all raw candidate anomalies generated in the first stage alongside the training data. The comprehensive ablation results are presented in TABLE~\ref{tab:ablation}.
\begin{table*}[t]
\centering
\caption{The evaluation results of ablation experiments.}
\setlength{\tabcolsep}{1mm}
\begin{tabular}{c|ccc|ccc|ccc|ccc|ccc}
\hline
\multirow{2}{*}{\textbf{Method/Dataset}} & \multicolumn{3}{c|}{\textbf{Wbc}} & \multicolumn{3}{c|}{\textbf{Pima}} & \multicolumn{3}{c|}{\textbf{Cardio}} & \multicolumn{3}{c}{\textbf{Yeast}} & \multicolumn{3}{c}{\textbf{Thyroid}}\\ 
\cline{2-16} 
 & \textbf{auc-roc} & \textbf{auc-pr} & \textbf{f1}
 & \textbf{auc-roc} & \textbf{auc-pr} & \textbf{f1}
 & \textbf{auc-roc} & \textbf{auc-pr} & \textbf{f1}
 & \textbf{auc-roc} & \textbf{auc-pr} & \textbf{f1}
 & \textbf{auc-roc} & \textbf{auc-pr} & \textbf{f1}\\
\hline

\textbf{\textit{w/o generation}} 
& 0.9347 & 0.8667 & \textbf{0.8128}
& 0.6989 & 0.7290 & \textbf{0.7035}
& 0.8071 & 0.6875 & 0.6642
& 0.5639 & 0.5528 & 0.6778
& 0.9504 & 0.7355 & 0.7303 \\

\textbf{\textit{w/o selection}} 
& 0.6422 & 0.3648 & 0.3999
& 0.6182 & 0.6567 & 0.6885
& 0.9347 & 0.7950 & 0.7387 
& 0.5549 & 0.5443 & 0.6737
& 0.9219 & 0.7185 & 0.6962 \\

\cdashline{1-16}

\textbf{PLAG} 
& 0.9710 & 0.8649 & 0.7550
& 0.6596 & 0.6838 & 0.7023
& 0.9695 & 0.8747 & \textbf{0.7964}
& 0.5982 & 0.5711 & \textbf{0.6807}
& 0.9796 & 0.8464 & \textbf{0.7986} \\
\hline
\end{tabular}
\label{tab:ablation}
\end{table*}

\begin{figure*}[t]
    \centering
    \includegraphics[width=\textwidth]{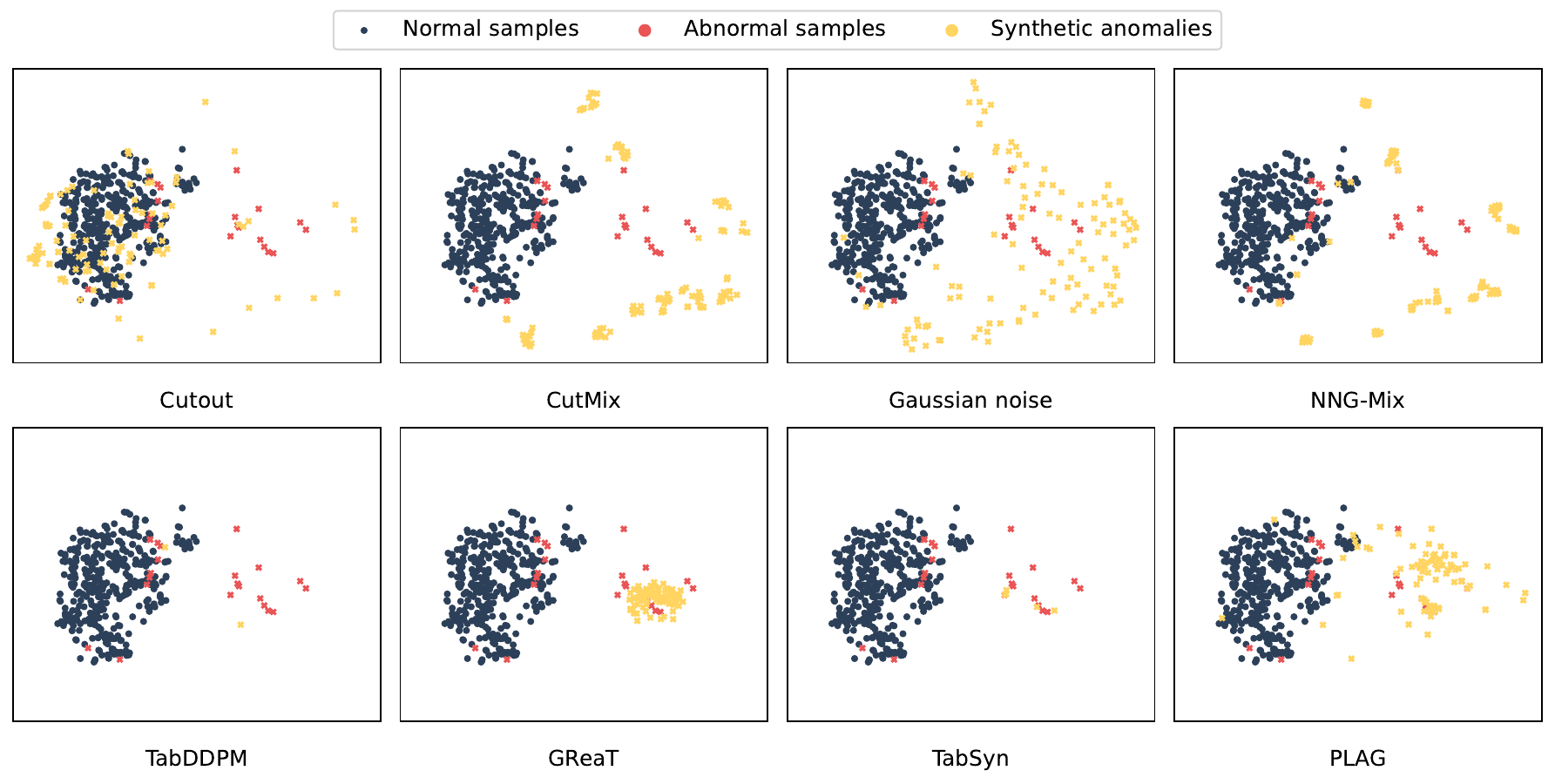}
    \caption{2D visualization comparing the distributions of synthetic anomalies generated by various methods against ground-truth normal and anomalous samples on the WBC dataset.}
    \label{fig:synthetic_anomaly_distri}
    \vspace{-1em}
\end{figure*}

A profound insight emerges from the detailed analysis of the table, although \textit{w/o generation} achieves optimal performance on the relatively smaller Wbc and Pima datasets, its efficacy degrades compared to the full PLAG framework as the dataset scale expands. 
We attribute this phenomenon to the inherent tension between data volume and the required diversity of anomaly patterns. 
In small-scale datasets, a limited number of highly representative pseudo-anomalies is often sufficient to delineate preliminary decision boundaries between normal and anomalous distributions. 
However, as the dataset volume substantially increases, these sparse initial pseudo-anomalies drastically fail to cover the intricate and diverse latent anomaly patterns, thereby bottlenecking the downstream performance.

This underlying logic is further corroborated by analyzing the \textit{w/o selection} variant. 
This variant similarly exhibits sub-optimal performance on smaller datasets but demonstrates a relatively positive trend on larger ones, indirectly confirming the critical demand for an increased volume of synthetic samples in massive datasets. 
Crucially, the full PLAG framework universally outperforms w/o selection across all metrics, providing irrefutable evidence for the necessity of filtering the candidate anomalies generated by the LLM. 
On the two largest datasets, Cardio and Thyroid, the two-stage filtering mechanism independently yields massive F1-score leaps of 0.0577 and 0.1024, respectively.

In conclusion, this ablation study not only substantiates the indispensability of each component but also yields a practical insight.
In real-world tabular anomaly detection, larger data volumes inherently necessitate a proportionally higher diversity of synthesized anomaly patterns. 
Concurrently, establishing a robust filtering mechanism to guarantee the quality of these generated instances serves as the fundamental cornerstone for shattering the performance ceiling of downstream classifiers.

\subsection{Visualization of Synthetic Anomalies}
To intuitively explain the performance improvements observed in the preceding quantitative evaluations, we project the synthetic anomalous samples generated by each method into a two-dimensional space. 
This allows for a comparative analysis against the underlying distributions of ground-truth normal and anomalous instances. 
It is important to note that SOEL is excluded from this visualization, as it merely generalizes anomaly scores to unlabeled samples without explicitly generating physical anomalous instances. 
Using the WBC dataset as a representative case study, the synthetic anomaly distributions produced by the remaining methods are illustrated in Fig.~\ref{fig:synthetic_anomaly_distri}.

Examining the generated distributions specifically, Cutout relies on randomly masking feature subsets to construct anomalies. 
This mechanism results in synthetic anomalous instances that heavily overlap with the normal sample clusters within the feature space. 
Such severe distributional overlap inevitably confounds the classifier's decision boundary, thereby explaining its poorest detection performance among all baselines. 
Conversely, CutMix attempts to synthesize anomalies by cutting and pasting features across different samples. 
This approach leads to an irregular and highly scattered distribution of anomalous representations in the latent space, which similarly poses substantial interference to the model's ability to discriminate between normal and anomalous boundaries. 
These visualizations collectively suggest that simply applying direct feature-level modifications (e.g., random masking or splicing) to original samples often fails to yield highly discriminative, realistic anomalous patterns. 
Consequently, this is usually ineffective as an anomaly synthesis strategy for complex tabular data scenarios.

Synthesizing anomalies by adding noise to pseudo-anomalous samples mimics the true anomaly distribution reasonably well, aiding the model in achieving respectable detection performance. 
However, its overall efficacy remains unstable. 
As indicated by the results in TABLE~\ref{tab:generation}, its detection performance falters significantly on certain datasets. 
This suggests that while noise-based anomaly generation can be viable in some cases, it is fundamentally not a stable or universally reliable synthesis strategy.

General-purpose tabular data synthesizers aim primarily to model the distribution of provided reference samples. 
Consequently, they exhibit a consistent generative pattern: synthesized anomalies cluster tightly around the initial pseudo-anomalies. 
Furthermore, the volume of anomalous instances generated by TabDDPM and TabSyn is severely restricted in the experiments. 
This is likely attributable to the scarcity of pseudo-anomalies available as references, indicating that such methods are ill-suited for tabular anomaly detection tasks lacking true labels.

In contrast, anomalies generated by PLAG not only align with typical true anomaly distributions but also moderately extend toward the boundaries of the normal class. 
This spatial characteristic effectively highlights the diversity of the synthesized anomalies and comprehensively delineates the latent anomalous manifold, ultimately yielding the best overall detection performance.

\section{Conclusion}
In this work, we proposed PLAG, a novel pseudo-label-guided anomaly generation framework tailored to enhance tabular anomaly detection. 
Specifically, by utilizing pseudo-anomalies as guidance signals, PLAG uniquely decouples the overall anomaly quantification of a sample into an accumulation of feature-level abnormalities. 
Guided by this fine-grained perspective, tailored prompt templates are constructed to direct an LLM in synthesizing candidate anomalies. 
Subsequently, a two-stage data selection strategy is introduced to filter these candidates, thereby guaranteeing the fidelity and diversity of them.
Extensive experiments across five datasets demonstrate that these filtered samples serve as robust discriminative guidance, empowering PLAG to achieve state-of-the-art performance against eight representative anomaly generation baselines. 
Furthermore, as a flexible framework, PLAG demonstrates broad applicability by seamlessly integrating with various existing unsupervised anomaly detectors.
Concurrently, our empirical results reveal a crucial insight that the necessity for synthetic anomaly diversity becomes increasingly critical as the dataset scale expands. 
In future work, we plan to develop an adaptive diversity control mechanism to enable personalized sample generation tailored to specific datasets. 
Additionally, exploring the zero-shot and few-shot reasoning capabilities of LLMs for tabular anomaly detection remains a pivotal direction for our ongoing research.

\bibliographystyle{IEEEtran}
\bibliography{reference}

\vfill

\end{document}